\documentclass[11pt,a4paper]{article}
\usepackage[hyperref]{emnlp2018}
\usepackage{times}
\usepackage[utf8]{inputenc}
\usepackage{latexsym}
\usepackage{amsmath}
\usepackage{amssymb}
\usepackage{color,colortbl}
\usepackage{booktabs}
\usepackage{subcaption}
\usepackage{multirow}
\usepackage{url}
\usepackage{ifthen}
\usepackage{xspace}
\usepackage{graphicx}
\usepackage{framed}
\usepackage{textcomp}
\usepackage{comment}
\usepackage{xcolor}

\aclfinalcopy 

\newif\ifcomment
\commentfalse

\newcommand\refsec[1]{Section~\ref{sec:#1}}

\newcommand\reffig[1]{Figure~\ref{fig:#1}}

\newcommand\reftab[1]{Table~\ref{tab:#1}}
\newcommand\refapp[1]{Appendix~\ref{sec:#1}}

\ifthenelse{\isundefined{\definition}}{\newtheorem{definition}{Definition}}{}
\ifthenelse{\isundefined{\assumption}}{\newtheorem{assumption}{Assumption}}{}
\ifthenelse{\isundefined{\hypothesis}}{\newtheorem{hypothesis}{Hypothesis}}{}
\ifthenelse{\isundefined{\proposition}}{\newtheorem{proposition}{Proposition}}{}
\ifthenelse{\isundefined{\theorem}}{\newtheorem{theorem}{Theorem}}{}
\ifthenelse{\isundefined{\lemma}}{\newtheorem{lemma}{Lemma}}{}
\ifthenelse{\isundefined{\corollary}}{\newtheorem{corollary}{Corollary}}{}
\ifthenelse{\isundefined{\alg}}{\newtheorem{alg}{Algorithm}}{}
\ifthenelse{\isundefined{\example}}{\newtheorem{example}{Example}}{}

\definecolor{darkgreen}{rgb}{0,0.5,0}
\ifcomment
\newcommand\pl[1]{\textcolor{red}{[PL: #1]}}
\newcommand\hh[1]{\textcolor{blue}{[HH: #1]}}
\newcommand\dc[1]{\textcolor{green}{[DC: #1]}}
\else
\newcommand\pl[1]{}
\newcommand\hh[1]{}
\newcommand\dc[1]{}

\fi

\newcommand{\sts}{sequence-to-sequence}

\newcommand{\crl}{Craigslist}
\newcommand{\cda}{coarse dialogue act}
\newcommand{\bow}{Bag-of-Words}

\newcommand{\cmss}[1]{{\fontfamily{cmss}\selectfont #1}}
\newcommand{\ut}[1]{\emph{``#1''}}
\newcommand{\lf}[1]{\cmss{#1}}

\newcommand{\cl}{\textsc{CraigslistBargain}}
\newcommand{\fb}{\textsc{DealOrNoDeal}}
\newcommand{\soc}{\textsc{SettlersOfCatan}}

\newcommand{\hbase}{Human}

\newcommand{\slword}{SL(word)}
\newcommand{\rlword}[1]{\rl{#1}(word)}
\newcommand{\slstate}{SL(act)}
\newcommand{\rlstate}[1]{\rl{#1}(act)}
\newcommand{\hybrid}{SL(act)+rule}
\newcommand{\rl}[1]{RL$_\text{#1}$}

\newcolumntype{L}[1]{>{\raggedright\let\newline\\\arraybackslash\hspace{0pt}}m{#1}}
\newcolumntype{C}[1]{>{\centering\let\newline\\\arraybackslash\hspace{0pt}}m{#1}}
\newcolumntype{R}[1]{>{\raggedleft\let\newline\\\arraybackslash\hspace{0pt}}m{#1}}

\usepackage{soul}

\title{Decoupling Strategy and Generation in Negotiation Dialogues}

\author{
    He He \and Derek Chen \and Anusha Balakrishnan \and Percy Liang\\
    Computer Science Department, Stanford University \\
    {\tt \{hehe,derekchen14,anusha,pliang\}@cs.stanford.edu}
}
\date{}

\begin{document}
\maketitle
\begin{abstract}
We consider negotiation settings
in which two agents use natural language to bargain on goods.
Agents need to decide on both high-level strategy (e.g., proposing \$50)
and the execution of that strategy (e.g., generating
\ut{The bike is brand new. Selling for just \$50!}).
Recent work on negotiation trains neural models,
but their end-to-end nature makes it hard to control their strategy,
and reinforcement learning tends to lead to degenerate solutions.
In this paper, we propose a modular approach based on \cda{}s (e.g., \lf{propose(price=50)})
that decouples strategy and generation.
We show that we can flexibly set the strategy
using supervised learning, reinforcement learning, or domain-specific knowledge
without degeneracy,
while our retrieval-based generation can maintain context-awareness and produce diverse utterances.
We test our approach on the recently proposed \fb{} game,
and we also collect a richer dataset based on real items on \crl{}.
Human evaluation shows that
our systems achieve higher task success rate
and more human-like negotiation behavior
than previous approaches.

\end{abstract}

\section{Introduction}
\label{sec:introduction}

A good negotiator needs to decide on
the \emph{strategy} for achieving a certain goal (e.g., proposing \$6000)
and the realization of that strategy via \emph{generation} of natural language (e.g.,
\ut{I really need a car so I can go to work, but all I have is 6000, any more and I won't be able to feed my children.}).

Most past work in NLP on negotiation focuses on strategy (dialogue management)
with either no natural language~\cite{cuayahuitl2015strategic,cao2018emergent}
or canned responses~\cite{keizer2017negotiation,traum2008multi}.
Recently, end-to-end neural models~\cite{lewis2017deal,he2017symmetric} are used to
simultaneously learn dialogue strategy and language realization
from human-human dialogues,
following the trend of using neural network models on both
goal-oriented dialogue~\cite{wen2017network,dhingra2017information}
and open-domain dialogue~\cite{sordoni2015neural,li2017adversarial,lowe2017ubuntu}.
However, these models have two problems:
(i) it is hard to control and interpret the strategies,
and (ii) directly optimizing the agent's goal through reinforcement learning
often leads to degenerate solutions where
the utterances become ungrammatical~\cite{lewis2017deal} or repetitive~\cite{li2016rl}.

\begin{figure*}[ht]
\centering
\includegraphics[width=2\columnwidth]{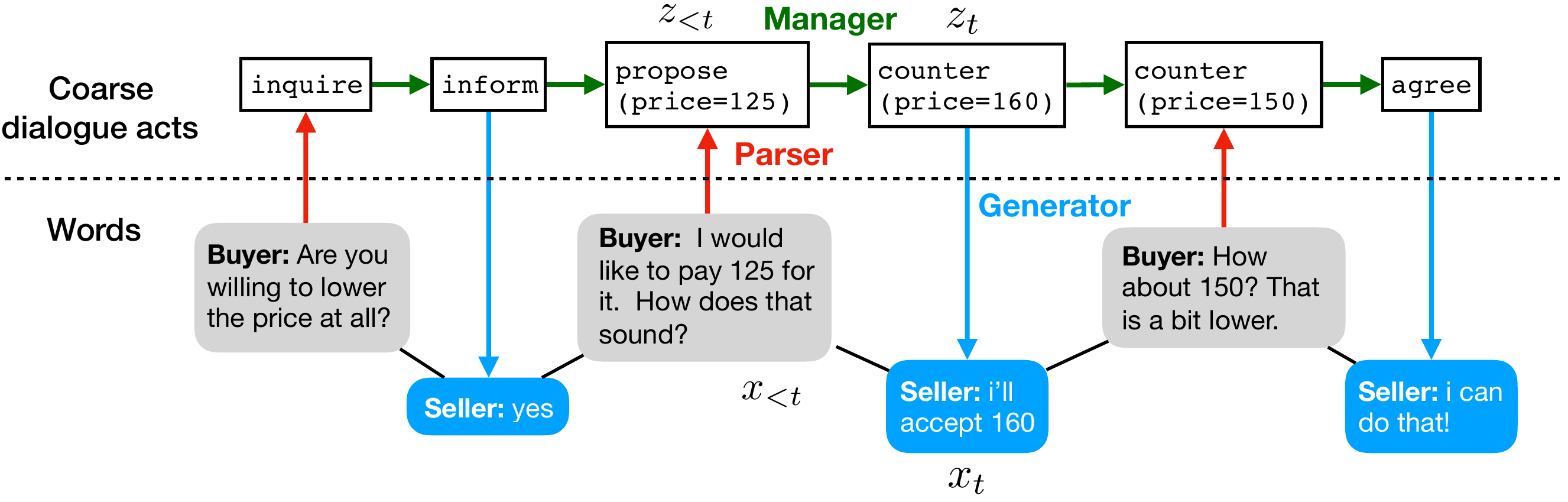}
\caption{Our modular framework consists of three components similar to traditional goal-oriented dialogue systems.
    (1) The parser maps received utterances to \cda{}s (an intent and its arguments)
    that capture the high-level dialogue flow.
    (2) The manager generates the next \cda{} $z_t$ conditioned on past dialogue acts $z_{<t}$.
    (3) The generator then produces a response
    conditioned on both the predicted \cda{} $z_t$ and the dialogue history $x_{<t}$.
    Importantly, unlike in traditional systems, \cda{}s only capture
    the rough shape of a dialogue, not
    the full meaning of its utterances,
    e.g., \lf{inform} does not specify the answer to the question.
  }
\label{fig:framework}
\end{figure*}

To alleviate these problems, our key idea is to
decouple strategy and generation, which
gives us control over the strategy such that
we can achieve different negotiation goals
(e.g., maximizing utility, achieving a fair deal)
with the same language generator.
Our framework consists of three components shown in \reffig{framework}:
First, the parser identifies keywords and entities to
map each utterance to a \emph{\cda{}}
capturing the high-level strategic move.
Then, the dialogue manager chooses a responding dialogue act
based on a \sts{} model over \cda{}s learned from parsed training dialogues.
Finally, the generator produces an utterance
given the dialogue act and the utterance history.

Our framework follows that of traditional goal-oriented dialogue systems \citep{young2013pomdp},
with one important difference:
\cda{}s are not intended to and cannot capture the full meaning of an utterance.
As negotiation dialogues are fairly open-ended, the generator needs to depend on the full utterance history.
For example, consider the first turn in \reffig{framework}.
We cannot generate a response given only the dialogue act \lf{inform}; we must also look at the previous question.
However, we still optimize the dialogue manager
in the \cda{} space using
supervised learning, reinforcement learning, or domain-specific knowledge.

Existing human-human negotiation datasets
are grounded in closed-domain games with a fixed set of objects
such as Settlers of Catan (lumber, coal, brick, wheat, and sheep)~\cite{afantenos2012modelling,asher2016catan}
or item division (book, hat, and ball)~\cite{devault2015toward,lewis2017deal}.
These objects lack the richness of the real world.
To study human negotiation in more open-ended settings
that involve real goods,
we scraped postings of items for sale from \url{craigslist.org} as our negotiation scenario.
By hiring workers on Amazon Mechanical Turk (AMT) to play the role of buyers and sellers,
we collected a new dataset (\cl{}) of negotiation dialogues.\footnote{
    Available at \url{https://stanfordnlp.github.io/cocoa}.
}
Compared to existing datasets,
our more realistic scenario
invites richer negotiation behavior involving open-ended aspects such as cheap talk or side offers.

We evaluate two families of systems modeling \cda{}s and words respectively,
which are optimized by supervised learning, reinforcement learning, or domain knowledge.
Each system is evaluated on our new \cl{} dataset and the \fb{} dataset of~\citet{lewis2017deal} by
asking AMT workers to chat with the system in an A/B testing setting.
We focus on two metrics: task-specific scores (e.g., utility) and human-likeness.
We show that reinforcement learning on \cda{}s avoids degenerate solutions,
which was a problem in \citet{li2016rl,lewis2017deal}.
Our modular model maintains reasonable human-like behavior while still optimizes the objective.
Furthermore, we find that models trained over \cda{}s
are stronger negotiators (even with only supervised learning)
and produce more diverse utterances
than models trained over words.
Finally, the interpretability of \cda{}s allows system developers to
combine the learned dialogue policy with hand-coded rules,
thus imposing stronger control over the desired strategy.

\begin{table*}[ht]
\begin{center}
\begin{minipage}[l]{0.29\textwidth}
    \centering
    {\footnotesize
    \caption*{\footnotesize{JVC HD-ILA 1080P 70 Inch TV}}
    \includegraphics[width=\textwidth]{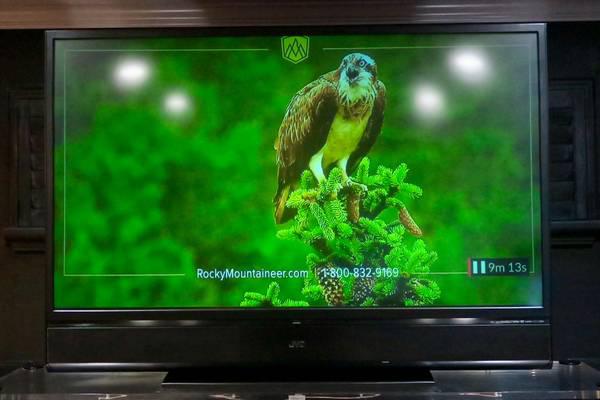}
    \caption*{
        \footnotesize{
            \\
            Tv is approximately 10 years old. Just installed new lamp. There are 2 HDMI inputs. Works and looks like new.\\
            Listing price: \$275 \\
            Buyer's target price: \$192
        }
    }
    }
\end{minipage}
\hfill
\begin{minipage}[c]{0.69\textwidth}
{\footnotesize
\definecolor{light-gray}{gray}{0.9}
\sethlcolor{light-gray}
\setlength\tabcolsep{0.5ex}
    \begin{tabular}{L{1cm}L{7.8cm}C{2cm}}
\toprule
Agent & Utterance & Dialogue Act \\
\midrule
        \rowcolor{light-gray}
Buyer & Hello do you still have the TV? & \lf{greet} \\

Seller & Hello, yes the TV is still available & \lf{greet} \\

        \rowcolor{light-gray}
Buyer & What condition is it in? Any scratches or problems? I see it recently got repaired & \lf{inquire} \\

Seller & It is in great condition and works like a champ! I just installed a new lamp in it. There aren't any scratches or problems. & \lf{inform} \\

        \rowcolor{light-gray}
Buyer & All right. Well I think 275 is a little high for a 10 year old TV.
        Can you lower the price some? How about 150? & \lf{propose(150)} \\ 

Seller & I am willing to lower the price, but \$150 is a little too low.
        How about \$245 and if you are not too far from me,
        I will deliver it to you for free?
        & \lf{counter(245)} \\

        \rowcolor{light-gray}
Buyer & It's still 10 years old and the technology is much older.
        Will you do 225 and you deliver it. How's that sound? 
        & \lf{counter(225)} \\

Seller & Okay, that sounds like a deal! & \lf{agree} \\

        \rowcolor{light-gray}
Buyer & Great thanks! & \lf{agree} \\

Seller & OFFER \$225.0 & \lf{offer(225)} \\

        \rowcolor{light-gray}
Buyer & ACCEPT & \lf{accept} \\
\bottomrule
\end{tabular}
}
\end{minipage}
\end{center}
\caption{Example dialogue between two people negotiating the price of a used TV.
}
\label{tab:cl-example}
\end{table*}

\section{Craigslist Negotiation Dataset}
\label{sec:craigslist}
Previous negotiation datasets were collected in the context of games.
For example,
\citet{asher2016catan} collected chat logs from online Settlers of Catan.
\citet{lewis2017deal} asked two people to
divide a set of hats, books, and balls.
While such games are convenient for grounding and evaluation,
it restricts the dialogue domain and the richness of the language.
Most utterances are direct offers such as \ut{has anyone got wood for me?}
and \ut{I want the ball.},
whereas real-world negotiation would involve more information gathering and persuasion.

To encourage more open-ended, realistic negotiation,
we propose the \cl{} task. 
Two agents are assigned the role of a buyer and a seller;
they are asked to negotiate the price of an item for sale on \crl{}
given a description and photos.
As with the real platform,
the listing price is shown to both agents.
We additionally suggest a private price
to the buyer as a target.
Agents chat freely in alternating turns.
Either agent can enter an offer price at any time, which can be
accepted or rejected by the partner.  Agents also have the option to quit,
in which case the task is completed with no agreement.

To generate the negotiation scenarios, we scraped postings on \url{sfbay.craigslist.org}
from the 6 most popular categories (housing, furniture, cars, bikes, phones, and electronics).
Each posting produces three scenarios with the buyer's target prices at
0.5x, 0.7x and 0.9x of the listing price.
Statistics of the scenarios are shown in \reftab{scenario-stats}.

We collected 6682 human-human dialogues
on AMT using the interface shown in \refapp{interface} \reffig{cl-website}.
The dataset statistics in \reftab{dataset-stats} show that
\cl{} has longer dialogues and more diverse utterances compared to prior datasets.
Furthermore,
workers were encouraged to embellish the item 
and negotiate side offers such as free delivery or pick-up.
This highly relatable scenario 
leads to richer dialogues such as the one shown in \reftab{cl-example}.
We also observed various persuasion techniques listed in \reftab{utterance-example}
such as embellishment, side offers, and appeals to sympathy.

\begin{table}[ht]
\centering
{\footnotesize
\begin{tabular}{lrr}
\toprule
    \# of unique postings & 1402 \\
    \% with images & 80.8 \\
    Avg \# of tokens per description & 42.6 \\
    Avg \# of tokens per title & 33.8 \\
    Vocab size & 12872 \\
\bottomrule
\end{tabular}
}
    \caption{Statistics of \cl{} scenarios.}
    \label{tab:scenario-stats}
\end{table}

\begin{table}[ht]
\centering
{\footnotesize
\begin{tabular}{lrrr}
\toprule
    & CB & DN & SoC\\
    \midrule
    \# of dialogues & 6682 & 5808 & 1081\\
    Avg \# of turns & 9.2 & 6.6 & 8.5\\
    Avg \# of tokens per turn & 15.5 & 7.6 & 4.2\\
    Vocab size & 13928 & 2719 & 4921\\
    Vocab size (excl. numbers) & 11799 & 2623 & 4735\\
\bottomrule
\end{tabular}
}
    \caption{Comparison of dataset statistics of \cl{} (CB),  \fb{} (DN), and 
    \soc{} (SoC).
    \cl{} contains longer, more diverse dialogues on average.}
    \label{tab:dataset-stats}
\end{table}

\begin{table*}[ht]
\centering
{\footnotesize
\setlength\tabcolsep{0ex}
\begin{tabular}{C{3cm}L{13cm}}
\toprule
Phenomenon & Example \\
\midrule
    Embellishment & It is in great condition and \textbf{works like a champ! I just installed a new lamp in it.} There aren't any scratches or problems. \\
    Cheap talk & How about i give you \$20 and you keep the helmet. \textbf{its for my daughter for her job, she delivers lemonade}. \\
    Side offers & \textbf{Throw in a couple of movies} with that DVD player, and you have yourself a deal. \\
    Appeal to sympathy & I would love to have this for my mother, \textbf{she is very sick} and this would help her and with me taking care of her and having to take a leave from work I can't pay very much of it \\
    World knowledge & For a \textbf{Beemer 5 series} in this condition, I really can't go that low. \\
\bottomrule
\end{tabular}
}
\caption{\label{tab:utterance-example}
    Rich negotiation language in our \cl{} dataset.
}
\end{table*}

\section{Approach}
\label{sec:approach}

\subsection{Motivation}

While end-to-end neural models have made promising progress in dialogue systems \cite{wen2017network,dhingra2017information},
we find they struggle to simultaneously learn
the strategy and the rich utterances necessary to succeed in the \cl{} domain,
e.g., \reftab{cl-human-bot-chats}(a) shows a typical dialogue
between a human and a \sts{}-based bot,
where the bot easily agrees.
We wish to now separate negotiation strategy and language generation.
Suppose the buyer says:
\ut{All right. Well I think 275 is a little high for a 10 year old TV.
    Can you lower the price some?  How about 150?}
We can capture the highest-order bit with a \cda{} \lf{propose(price=150)}.
Then, to generate the seller's response, the agent can first focus on this \cda{} rather than having to ingest the free-form text all at once.
Once a counter price is decided,
the rest is open-ended justification for the proposed price,
e.g., emphasizing the quality of the TV despite its age.

Motivated by these observations,
we now describe a modular framework
that extracts \cda{}s from utterances,
learns to optimize strategy in the dialogue act space,
and uses retrieval to fill in the open-ended parts conditioned on the full dialogue history.

\subsection{Overview}

Our goal is to build a dialogue agent that takes the dialogue history,
i.e. a sequence of utterances $x_1, \ldots, x_{t-1}$ along with
the dialogue scenario $c$ (e.g., item description),
and produces a distribution over the responding utterance $x_{t}$.

For each utterance $x_t$ (e.g., \ut{I am willing to pay \$15}), we define a \cda{} $z_t$ (e.g., \lf{propose(price=15)});
the \cda{} serves as a logical skeleton which does not
attempt to capture the full semantics of the utterance.
Following the strategy of traditional goal-oriented dialogue systems \citep{young2013pomdp},
we broadly define our model in terms of the following three modules:
\begin{enumerate}
  \itemsep0em
  \item A \textbf{parser} that (deterministically) maps an input utterance $x_{t-1}$
      into a \cda{} $z_{t-1}$ given the dialogue history $x_{<t}$ and $z_{<t}$,
      as well as the scenario $c$.
  \item A \textbf{manager} that predicts the responding dialogue act $z_t$ given past \cda{}s $z_{<t}$ and the scenario $c$.
  \item A \textbf{generator} that turns the \cda{} $z_t$
  to a natural language response $x_t$ given the full dialogue history $x_{<t}$.
\end{enumerate}
Because \cda{}s do not capture the full semantics,
the parser and the generator maintains full access to the dialogue history.
The main restriction is the manager examining the dialogue acts,
which we show will reduce the risk of degeneracy during reinforcement learning \refsec{human-eval}.
We now describe each module in detail (\reffig{framework}).

\subsection{Parser}
Our framework is centered around the \cda{} $z$,
which consists of an intent and a set of arguments.
For example, \ut{I am willing to pay \$15}
is mapped to \lf{propose(price=15)}.
The fact that our \cda{}s do not intend to
capture the full semantics of a sentence
allows us to use a simple rule-based parser.
It detects the intent and its arguments by regular expression matching
and a few if-then rules.
Our parser starts by detecting entities (e.g., prices, objects) and matching keyword patterns (e.g., \ut{go lower}).
These signals are checked against an ordered list of rules,
where we choose the first matched intent in the case of multiple matches.
An \lf{unknown} act is output if no rule is triggered.
The list of intent parsing rules used
are shown in \reftab{intent-detection}.
Please refer to \refapp{app-parser} for argument parsing based on entity detection.

\begin{table}[t]
\centering
{\footnotesize
\setlength\tabcolsep{0.3ex}
\begin{tabular}{C{1.9cm}L{5.6cm}}
\toprule
\multicolumn{2}{l}{ \textbf{Generic Rules} }                        \\
\midrule
\midrule
Intent        & Matching Patterns                                   \\
\midrule
\lf{greet}    & \textit{hi, hello, hey, hiya, howdy}                \\
\lf{disagree} & \textit{no, not, n't, nothing, dont}                \\
\lf{agree}    & not \lf{disagree} and \textit{ok, okay, great, perfect, deal, that works, i can do that}  \\
\lf{insist}   & the same offer as the previous one is detected \\
\lf{inquire}  & starts with an interrogative word (e.g., \textit{what, when, where}) or particle (e.g., \textit{do, are}) \\

\toprule
\multicolumn{2}{l}{ \textbf{\cl{} Rules} }                 \\
\midrule
\midrule
Intent        & Matching Patterns                                   \\
\midrule
\lf{intro}    & \lf{greet} or \textit{how are you, interested} \\
\lf{propose}  &  first price mention          \\
\lf{vague-price}  & no price mention and \textit{come down, highest, lowest, go higher/lower, too high/low}    \\
\lf{counter}  & new price detected     \\
\lf{inform}   & previous \cda{} was \lf{inquire}                       \\
\toprule
\multicolumn{2}{l}{ \textbf{\fb{} Rules} }                 \\
\midrule
\midrule
Intent        & Matching Patterns                                   \\
\midrule
\lf{propose}  & items and respective counts are detected               \\
\bottomrule
\end{tabular}
}
  \caption{Rules for intent detection in the parser.
  }
\label{tab:intent-detection}
\end{table}



\subsection{Manager}
\label{sec:manager}
The dialogue manager decides what action $z_t$
the dialogue agent should take at each time step $t$
given the sequence of past \cda{}s $z_{<t}$
and the scenario $c$.
Below, we describe three ways to learn the dialogue manager
with increasing controllability:
modeling human behavior in the training corpus (supervised learning),
explicitly optimizing a reward function (reinforcement learning),
and injecting hand-coded rules (hybrid policy).

\paragraph{Supervised learning.}
Given a parsed training corpus,
each training example is a sequence of \cda{}s over one dialogue,
$z_1, \ldots, z_T$.
We learn the transition probabilities $p_\theta(z_{t} \mid z_{<t}, c)$
by maximizing the likelihood of the training data. 

We use a standard \sts{} model with attention.
Each \cda{} is represented as a sequence of tokens,
i.e. an intent followed by each of its arguments, e.g., ``\lf{offer 150}''.
During the agent's listening turn,
an LSTM encodes the received \cda{};
during its speaking turn,
another LSTM decodes the tokens in the \cda{}.
The hidden states are carried over the entire dialogue to provide full history.

The vocabulary of \cda{}s is much smaller than the word vocabulary.
For example, our implementation includes fewer than 10 intents
and argument values are normalized and binned (see \refsec{model-implementation}).

\paragraph{Reinforcement learning.}
Supervised learning aims to mimic the average human behavior, but sometimes we want to directly optimize for a particular dialogue goal.
In reinforcement learning,
we define a reward $R(z_{1:T})$ on the entire sequence of \cda{}s.
Specifically, we experiment with three reward functions:
\begin{itemize}
    \item\textbf{Utility} is the objective of a self-interested agent.
        For \cl{}, we set the utility function to be a linear function of the final price,
        such that
        the buyer has a utility of 1 at their target price,
        the seller has a utility of 1 at the listing price,
        and both agents have a utility of zero at the midpoint of the listing price and the buyer's target price,
        making it a zero-sum game.
        For \fb{}, utility is the total value of objects given to the agent.
    \item \textbf{Fairness} aims to achieve equal outcome for both agents,
        i.e. the difference between two agents' utilities.
    \item \textbf{Length} is the number of utterances in a dialogue, thus
        encourages agents to chat as long as possible.
\end{itemize}
The reward is $-1$ if no agreement is reached.

We use policy gradient~\cite{williams1992simple} for optimization.
Given a sampled trajectory $z_{1:T}$ and the final reward $r$,
let $a_i$ be the $i$-th generated token (i.e. ``action'' taken by the policy) along the trajectory.
We update the parameters $\theta$
by 
\begin{equation}
\theta \leftarrow \theta - \eta \sum_i \nabla_\theta \log p_\theta(a_i \mid a_{<i}, c) (r - b)
\end{equation}
where $\eta$ is the learning rate
and $b$ is a baseline estimated by the average return so far for variance reduction.

\paragraph{Hybrid policy.}
Given the interpretable \cda{}s, a simple option is to
write a rule-based manager with domain knowledge,
e.g., if $z_{t-1}=\text{\lf{greet}}$, then $z_t=\text{\lf{greet}}$.
We combine these rules with a learned manager to fine-tune the dialogue policy.
Specifically, the dialogue manager predicts the intent
from a learned sequence model
but fills in the arguments (e.g., \lf{price}) using rules.
For example, given a predicted intent \lf{propose},
we can set the price to be the average of the buyer's and seller's current proposals
(a split-the-difference strategy).

\subsection{Generator}
We use retrieval-based generation to condition on both the \cda{} and the dialogue history.
Each candidate in our database for retrieval
is a tuple of an utterance $x_t$ and its dialogue context $x_{t-1}$,
represented by both templates and \cda{}s.
i.e. $(d(x_{t-1}), z_{t-1}, d(x_t), z_t)$,
where $d$ is the template extractor.
Specifically, given a parsed training set, each utterance is converted to a template by delexicalizing arguments in its \cda{}.
For example, \ut{How about \$150?} becomes ``How about \cmss{[price]}?'',
where \cmss{[price]} is a placeholder to be filled in at generation time.

At test time,
given $z_t$ from the dialogue manager,
the generator first retrieves candidates with the same intent as $z_t$ and $z_{t-1}$.
Next, candidates are ranked by similarity between their context templates and the current dialogue context.
Specifically, we represent the context $d(x_{t-1})$ as a TF-IDF weighted bag-of-words vector
and similarity is computed by a dot product of two context vectors.
To encourage diversity, the generator samples an utterance from the top $K$ candidates
according to the distribution given by
a trigram language model estimated on the training data.

\section{Experiments}
\label{sec:experiments}

\subsection{Tasks}
We test our approach on two negotiation tasks.
\textbf{\cl{}} (\refsec{craigslist}) asks a buyer and a seller to
        negotiate the price of an item for sale given its Craigslist post.
\textbf{\fb{}}~\cite{lewis2017deal} asks two agents to
        divide a set of items given their private utility functions.

\subsection{Models}
\label{sec:model-implementation}
We compare two families of models:
end-to-end neural models that directly map the input dialogue context
to a sequence of output words,
and our modular models that use \cda{}s as the intermediate representation.

We start by training the word-based model and the act-based model with supervised learning (SL).
\begin{itemize}
    \itemsep0em
    \item \textbf{\slword{}}: a \sts{} model with attention over
        previous utterances and the scenario,
        both embedded as a continuous \bow{};
    \item \textbf{\slstate{}}: our model described in \refsec{approach}
        with a rule-based parser, a learned neural dialogue manager, and a retrieval-based generator.
\end{itemize}

To handle the large range of argument values (prices) in \cl{} for act-based models,
we normalize the prices such that an agent's target price is 1
and the bottomline price is 0.
For the buyer, the target is given and the bottomline is the listing price.
For the seller, the target is the listing price and the bottomline is set to 0.7x of the listing price.
The prices are then binned according to their approximate values with two digits after the decimal point.

Next, given the pretrained SL models,
we fine-tune them with the three reward functions (\refsec{manager}),
producing \textbf{\rl{utility}}, \textbf{\rl{fairness}},
and \textbf{\rl{length}}. 

In addition, we compare with the hybrid model, \textbf{\hybrid{}}.
It predicts the next intent using a trigram language model learned over intent sequences in the training data,
and fills in the arguments with hand-coded rules.
For \cl{}, the only argument is the price.
The agent always splits the difference when making counter proposals,
rejects an offer if it is worse than its bottomline
and accepts otherwise.
For \fb{}, the agent maintains an estimate of the partner's private utility function.
In case of disagreement, it gives up the item with the lowest value of
$(\text{own utility} - \text{partner utility})$
and takes an item of estimated zero utility to the partner.
The agent agrees whenever a proposal is better than the last one
or its predefined target.
A high-level comparison of all models is shown in \reftab{model-comparison}.

\begin{table}[t]
\centering
{\footnotesize
\setlength\tabcolsep{3pt}
\begin{tabular}{lcccc}
\toprule
\textbf{Model}  &  $z$ & \textbf{Parser}
                & \textbf{Manager}               & \textbf{Generator}        \\
\midrule
SL/\rlword{}      &   vector    &  learned  &   learned    & generative \\
SL/\rlstate{}     & logical &   rules   &   learned    & retrieval  \\
\hybrid{}         & logical &   rules   &    hybrid    & retrieval  \\
\bottomrule
\end{tabular}
}
    \caption{Comparison of different implementation of the core modules in our framework.}
\label{tab:model-comparison}
\end{table}

\begin{table*}[t]
\centering
\setlength\tabcolsep{5pt}
\definecolor{light-gray}{gray}{0.9}
\sethlcolor{light-gray}
\begin{tabular}{L{2.6cm}rrrrr|rrrrr}
\toprule
                        & \multicolumn{5}{c|}{\cl{}} & \multicolumn{5}{c}{\fb{}} \\
                        &   Hu    &    Ut    &     Fa    &   Ag    &  Len    
                        &   Hu   &    Ut           &     Fa   &  Ag    &  Len    \ \\
\midrule                                                                                                                               
\midrule                                                                                                                               
    \hbase{}            &   4.3   &    -0.07  &   -0.14   &  0.91   &  10.2   
                        &  4.6   &  5.5 vs. 5.3     &    -0.2   &  0.78  &   5.8   \\
\midrule                                                                                                                               
    \slword{}           &   3.0   &   -0.32  &   -0.64   &  0.75   &  7.8    
                        &  \bf{3.8}   &  4.7 vs. 5.0     &    -0.3   &  0.70  &   5.0   \\

    \slstate{}          &   3.3   &   0.06   &   -0.12   &  0.84   &  14.0   
                        &  3.2   &  5.2 vs. 5.0     &    -0.2   &  0.67  &   7.0   \\

    \hybrid{}           & \bf{3.6}&   0.23   &    -0.46  &  0.75   &  11.4   
                        &  \bf{4.2}   &  5.2 vs. 5.2     &   0   &  0.72  &   8.0   \\
\midrule                                                                                                                               
    \rlword{utility}    & 1.7     & \hl{1.00}&  -2.00    &  0.31   & 2.5     
                        &  1.7   &\hl{2.9 vs. 1.8}  &    -1.1   &  0.33  &  10.4   \\

    \rlstate{utility}   & \bf{2.8} & \hl{1.00}&-2.00      &  0.22   & 6.7     
                        & \bf{2.8}   &\hl{3.3 vs. 2.3}  &    -1.0   &  0.38  &   9.5   \\
\midrule                                                                                                                               
    \rlword{fairness}   &   1.8   &   -0.62  & \hl{-1.24}&  0.75   &  9.4    
                        &  3.2   &  5.7 vs. 5.9     & \hl{-0.2} &  0.79  &   4.0   \\

    \rlstate{fairness}  & \bf{3.0}&   -0.28  & \hl{-0.56}&  0.68   &  7.1    
                        &  3.5   &  4.2 vs. 5.4     & \hl{-1.2} &  0.77  &   7.6   \\
\midrule                                                                                                                               
    \rlword{length}     & 1.9     &   -0.79  &  -1.58    &  0.85   &\hl{13.8}
                        &  1.6   &  3.4 vs. 2.9     &    -0.5   &  0.48  &\hl{9.2} \\

    \rlstate{length}    & \bf{3.0}&   0.89   &  -1.78    &  0.40   &\hl{11.8}
                        &  \bf{2.5}   &  2.5 vs. 3.1     &  -0.6     &  0.54  &\hl{11.0} \\
\bottomrule
\end{tabular}
\caption{
    Human evaluation results on human-likeness (Hu), agreement rate (Ag),
    and RL objectives,
    including agent utility (Ut),
    deal fairness (Fa), and dialogue length (Len).
    Results are grouped by the optimization objective.
    For each group of RL models, the column of the optimization objective is \hl{highlighted}.
    For human-likeness, scores that are better than others in the same group with statistical significance
    ($p<0.05$ given by paired $t$-tests) are in \textbf{bold}.
    Overall, with SL, all models are human-like, however, act-based models better matches human statistics across all metrics;
    with RL, word-based models becomes degenerate, whereas act-based models optimize the reward while maintaining human-likeness.
    }
\label{tab:human-eval}
\end{table*}

\renewcommand{\bot}[1]{{\bf #1}}
\begin{table*}[ht]
{\footnotesize
\setlength\tabcolsep{0.5ex}
\begin{tabular}{p{8cm}l}
\toprule
\textbf{Title:} Fitbit Flex + 2 Bands Activity Tracker   & \textbf{Listing Price:} \$60  \\
\multicolumn{2}{l}{\textbf{Description:} Hi, I'm selling a Fitbit Activity Tracker. Track your day including steps, distance, calories burned, active minutes.} \\
\multicolumn{2}{l}{Track your night including hours slept, times to woken up, sleep quality, Wake with a silent vibrating alarm. You can set a ...} \\ 
\midrule
\midrule
    (a) A: \slword{} (Buyer) \quad B: Human (Seller) &
    (b) A: \slstate{} (Buyer)  \quad B: Human (Seller) \\
\midrule
    \begin{tabular}[t]{l}
        A: \bot{hi, i 'm interested in your fitbit.} \\
        B: Great. It really is a good deal. \\
        A: \bot{great.}  \\ B: Do you have any questions about it? \\
        A: \bot{i 'm interested in your fitbit. i 'm not sure i 'm not sure} \\
        \bot{i 'm interested.}  \\
        B: It has so many functions that are great. I particularly like \\
        the sleep tracking. Plus it is waterproof. How about I will \\
        deliver it to you and you give me 60 dollars? \\
        A: \bot{i can do \$52 if you come pick it up.}  \\
        B: Well I am selling it. So I will deliver it for 60. Deal? \\
        A: \bot{yes, i can do that.}   \\
        B: OFFER(\$60) \qquad   A: \bot{ACCEPT}
    \end{tabular} &
    \begin{tabular}[t]{l}
        A: \bot{how we doing today} \\
        B: Hi!\\
        A: \bot{yes, i really need one, what color is it?} \\
        B: It has a black band and a white band.\\
        A: \bot{i see, so you bought an extra one by mistake?} \\
        B: No, it comes with 2 bands \\
        A: \bot{i would like to make you a offer of 36} \\
        B: Ooooh....yeah, I can't. Sorry. It's wireless, has Bluetooth, \\
         extra bands, and is water resistant. I could do \$50.\\
        A: \bot{49?} \\
        B: You know..I'd do \$49. \\
        A: \bot{thank you so much!} \\
        B: OFFER(\$49) \qquad   A: \bot{ACCEPT}
    \end{tabular} \\
\midrule
\midrule
    (c) A: \rlword{length} (Buyer) \quad B: Human (Seller) &
    (d) A: \rlstate{length} (Buyer)  \quad B: Human (Seller) \\
\midrule
    \begin{tabular}[t]{l}
        A: \bot{hello} \\
        B: Hi how are you?\\ 
        A: \bot{i am.}  \\
        B: you are interested in the fitbit flex and the 2 trackers? \\
        A: \bot{it's in good condition condition.} \\
        B: yes \\
        A: \bot{ok, that sounds good.} \\
        B: I am selling it at \$60? \\
        A: \bot{it's in good condition condition.} \\
        B: yes \\
        A: \bot{ok, that sounds good.} \\
        B: \$60 \\
        A: \bot{i can go that low.}
    \end{tabular} &
    \begin{tabular}[t]{l}
        A: \bot{hi, i saw your ad about the item.} \\
        B: Okay great, i'm selling a Fitbit Flex plus 2 bands \\
        A: \bot{how old is the item?}  \\
        B: I've only had it for about a few months. \\
        A: \bot{but it does work correct?} \\
        B: yes it does work, it's in great condition. \\ 
        A: \bot{excellent. would you consider taking 36 for it?}  \\
        B: I'm selling for \$60 but \$36 would be way too low. \\ 
        A: \bot{how about 36?}   \\
        B: no I cannot accept \$36, I'm sorry \\
        A: \bot{i'm sorry. would you accept 36?}   \\
        B: I am not going lower than \$50 \\ 
        A: \bot{OFFER(\$36)} \qquad   A: REJECT 
    \end{tabular} \\
\bottomrule
\end{tabular}
} 
\caption{Example human-bot chats on \cl{}, where bot utterances are in bold.
\slword{} produced generic responses, while \slstate{} is more human-like.
\rlword{length} devolved into degenerate behavior repeating itself while \rlstate{length} maintained coherency.
Only the first half of the item description and the \rlword{length} chat are shown due to space limit.
}
\label{tab:cl-human-bot-chats}
\end{table*}

\subsection{Training Details}

\paragraph{\cl{}}
For \slword{},
we use a \sts{} model with attention over 3 previous utterances
and the negotiation scenario (embedded as a continuous \bow{}).
For both \slword{} and \slstate{},
we use 300-dimensional word vectors initialized by pretrained GloVe word vectors~\cite{pennington2014glove},
and a two-layer LSTM with 300 hidden units for both the encoder and the decoder.
Parameters are initialized by sampling from a uniform distribution between -0.1 and 0.1.
For optimization, we use AdaGrad~\cite{duchi10adagrad}
with a learning rate of 0.01 and a mini-batch size of 128.
We train the model for 20 epochs and
choose the model with the lowest validation loss.

For RL, we first fit a partner model using supervised learning (e.g., \slword{}),
then run RL against it.
One agent is updated by policy gradient and the partner model is fixed during training.
We use a learning rate of 0.001 and train for 5000 episodes (dialogues).
The model with the highest reward on the validation set is chosen.

\paragraph{\fb{}}
For act-based models,
we use the same parameterization as \cl{}.
For word-based models,
we use the implementation from \citet{lewis2017deal}.\footnote{\url{https://github.com/facebookresearch/end-to-end-negotiator}}
Note that for fair comparison, we did not apply SL interleaving during RL training and rollouts during inference.

\subsection{Human Evaluation}
\label{sec:human-eval}
We evaluated each system on two metrics:
task-specific scores (e.g., utility) and human-likeness.
The scores tell us how well the system is playing the game, 
and human-likeness tells us whether the bot deviates from human behavior,
presumably due to over-optimization.

We put up all 9 systems online and
hired workers from AMT to chat with the bots.
Each worker was randomly paired with one of the bots or another worker,
so as to compare the bots with human performance under the same conditions.
At the end of a chat, workers were asked the question
\ut{Do you think your partner demonstrated reasonable human behavior?}.
They provided answers 
on a Likert scale from 1 (not at all) to 5 (definitely).
\reftab{human-eval}
shows the human evaluation results on \cl{} and \fb{} respectively. 
We also show example human-bot dialogues in \reftab{cl-human-bot-chats}
and \refapp{human-bot-chats}.

\paragraph{\slstate{} learns more human-like behavior.}
We first compare performance of SL models over words and \cda{}s.
Both \slword{} and \slstate{} achieved similar scores on human-likeness
(no statistically significant difference).
However, \slword{} better matched human statistics such as dialogue length and utility. 
For instance, \slword{} tended to produce short, generic utterances as shown in \reftab{cl-human-bot-chats}(a); 
they also agreed on a deal more quickly because
utterances such as \ut{deal} and \ut{I can do that} are frequent in negotiation dialogues.
This behavior is reflected by the shorter dialogue length and lower utility of \slword{} models.

\paragraph{\rlword{} leads to degeneracy.}
On \cl{},
all \rlword{} models clearly have low scores on human-likeness in \reftab{human-eval}.
They merely learned to repeat a few sentences:
The three most frequent sentences 
of \rlword{utility}, \rlword{fairness}, and \rlword{length} account for 81.6\%, 100\% and 100\% of all utterances.
For example, \rlword{utility} almost always opened with \ut{i can pick it up},
then offer its target price.
\rlword{length} repeated generic sentences until the partner submitted a price.
While they scored high on the reward being optimized,
the conversations are unnatural.

On \fb{}, we have observed similar patterns.
A general strategy learned by \rlword{} was to pick an offer depending on its objective,
then repeat the same utterance over and over again
(e.g., \ut{i need the ball.}), resulting in low human-likeness scores.
One exception is \rlword{fairness}, since most of its offers were reasonable and
agreed on immediately (it has the shorted dialogue length),
the conversations are natural.

\paragraph{\rlstate{} optimizes different negotiation goals while being human-like.}
On both tasks, \rlstate{} models optimized their rewards while maintaining reasonable human-likeness scores.
We now show that different models demonstrated different negotiation behavior.
Two main strategies learned by \rlstate{length}
were to ask questions and to postpone offer submission.  
On \cl{}, when acting as a buyer, 42.4\% of its utterances were questions,
compared to 30.2\% for other models.
On both tasks, it tended to wait for the partner to submit an offer (even after a deal was agreed on), 
compared to \rlstate{margin} which almost always submitted offers first.
For \rlstate{fairness}, it aimed to agree on a price in the middle of the listing price and the buyer's target price for \cl{}.
Since the buyer's target was hidden,
when the agent was the seller, it tended to wait for the buyer to propose prices first.
Similary, on \fb{} it waited to hear the parter's offer and sometimes changed its offer afterwards,
whereas the other models often insisted on one offer.

On both tasks, \rlstate{utility} 
learned to insist on its offer and refuse to budge.
This ended up frustrating many people, which is why it has a low agreement rate.
The problem is that our human model is simply a SL model trained on human-human dialogues,
which may not accurately reflects real human behavior during human-bot chat.
For example, the SL model often agrees after a few turns of insistence on a proposal,
whereas humans get annoyed if the partner is not willing to make compromises at all.
However, by injecting domain knowledge to \hybrid{},
e.g., making a small compromise is better than stubbornly being fixed on a single price,
we were able to achieve high utility and human-likeness on both \cl{} and \fb{}.

\section{Related Work and Discussion}
\label{sec:discussion}

Recent work has explored the space between
goal-oriented dialogue and open-domain chit-chat
through collaborative or competitive language games,
such as
collecting cards in a maze~\cite{potts2012cards},
finding a mutual friend~\cite{he2017symmetric},
or splitting a set of items~\cite{devault2015toward,lewis2017deal}.
Our \cl{} dialogue falls in this category,
but exhibits richer and more diverse language than prior datasets.
Our dataset calls for systems that can handle both strategic decision-making 
and open-ended text generation. 

Traditional goal-oriented dialogue systems build a pipeline of modules
\cite{young2013pomdp,williams2016dstc}.
Due to the laborious dialogue state design and annotation,
recent work has been exploring ways to replace these modules with neural networks and end-to-end training
while still having a logical backbone~\cite{wen2017network,bordes2017learning,he2017symmetric}.
Our work is closely related to the Hybrid Code Network~\cite{williams2017dialog},
but the key difference is that \citet{williams2017dialog}
uses a neural dialogue state,
whereas we keep a structured, interpretable dialogue state
which allows for stronger top-down control.
Another line of work tackles this problem by
introducing latent stochastic variables to model the dialogue state~\cite{wen2017latent,zhao2017learning,cao2017latent}.
While the latent discrete variable allows for post-hoc discovery of dialogue acts 
and increased utterance diversity,
it does not provide controllability over the dialogue strategy.

Our work is also related to a large body of literature on dialogue policies
in negotiation~\cite{english2005mixed,efstathiou2014catan,hiraoka2015trading,cao2018emergent}.
These work mostly focus on learning good negotiation policies in a domain-specific action space,
whereas our model operates in an open-ended space of natural language.  
An interesting future direction is to connect with
game theory~\cite{brams2003negotiation} for complex multi-issue bargaining.
Another direction is learning to generate persuasive utterances,
e.g., through framing~\cite{hiraoka2014framing} or
accounting for the social and cultural context~\cite{nouri2012cultural}.

To conclude, we have introduced \cl{},
a rich dataset of human-human negotiation dialogues.
We have also presented a modular approach based on coarse dialogue acts
that models a rough strategic backbone as well allowing for open-ended generation.
We hope this work will spur more research in hybrid approaches
that can work in open-ended, goal-oriented settings.

\paragraph{Acknowledgments.}
This work is supported by DARPA Communicating with Computers (CwC)
program under ARO prime contract no. W911NF-15-1-0462. 
We thank members of the Stanford NLP group for insightful discussion
and the anonymous reviewers for constructive feedback.

\paragraph{Reproducibility.}
All code, data, and experiments for this paper are
available on the CodaLab platform:
{\footnotesize \url{https://worksheets.codalab.org/worksheets/0x453913e76b65495d8b9730d41c7e0a0c/}}.

\bibliography{all}
\bibliographystyle{acl_natbib_nourl}

\clearpage
\appendix
\begin{figure*}[ht]
\centering
\includegraphics[width=\textwidth]{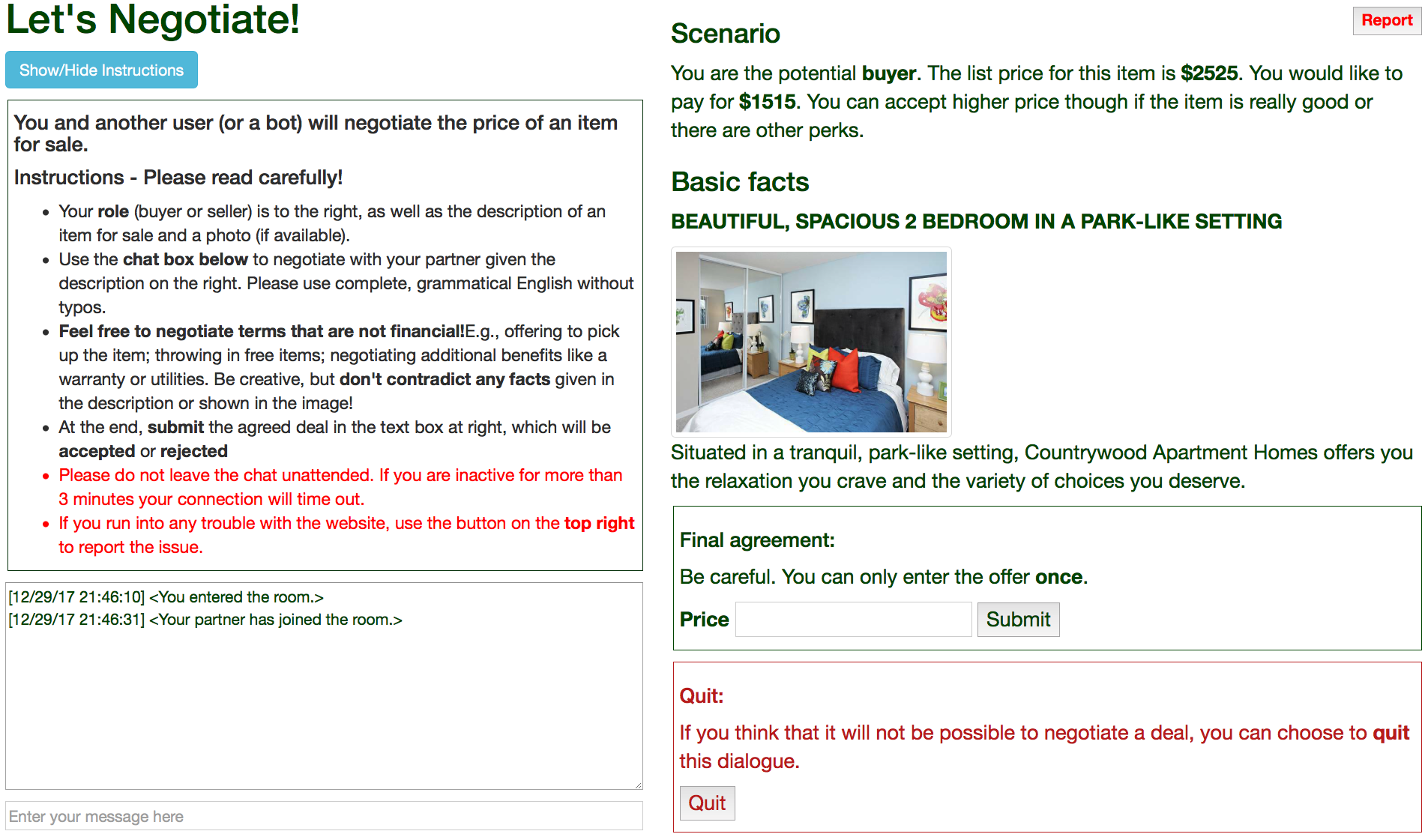}
\caption{Our chat interface for \cl{}.}
\label{fig:cl-website}
\end{figure*}

\section{\cl{} Web Interface}
\label{sec:interface}

\reffig{cl-website} shows our web interface where workers negotiate.

\section{Argument Detection of the Rule-based Parser}
\label{sec:app-parser}

\paragraph{Price detection.}
On \cl{}, given an utterance, we want to detect mentioned prices in it,
which are arguments of intents such as \lf{propose} and \lf{counter}.
We first detect groud truth prices in the training data, which are numbers starting or ending with the dollar sign.
At test time, a number is considered a price if
it starts or ends with the dollar sign,
or (a) its left and right neighboring words appear next to ground truth prices in the training data
and (b) it is not larger than 1.5x of the listing price.

\paragraph{Item and count detection.}
On \fb{}, given an utterance, we want to parse the proposed split of items,
i.e. numbers of balls, hats, and books for each agent.
We first detect first/second person pronouns, the three objects (ball, hat, and book), and counts (1 to 10) by regular expression matching.
To decide the grouping of agent, object, and count,
we process the utterance from left to right;
as soon as a pair of object and count is detected, we group it with the most recently referred agent by resolving the pronouns (e.g., \ut{I} or \ut{you}).

\section{Example Dialogues}
\label{sec:human-bot-chats}
Examples of human-bot chats on \fb{} are shown in \reftab{fb-human-bot-chats}, where bot utterances are in bold.
The full set of evaluation dialogues are available on the Codalab worksheet.

\renewcommand{\bot}[1]{{\bf #1}}
\begin{table*}[ht]
{\footnotesize
\centering
\begin{tabular}{lccc}
\toprule
       & Item Count & Value to Bot & Value to Human \\
\midrule
Book:  &     2      &      4       &        3       \\
Hat:   &     1      &      2       &        0       \\
Ball:  &     2      &      0       &        2       \\
\end{tabular} \\
\begin{tabular}{ll}
\midrule
\midrule
A: \slword{} \quad B: Human &   A: \slstate{} \quad B: Human         \\
\midrule
    \begin{tabular}[t]{l}
        A: \bot{i would like the books and the hat .}                \\
        B: you can have 1 book and a ball , i need a hat and a book  \\
        A: \bot{ok}                                                  \\
        B: deal                                                      \\
        A: \bot{DEAL AGREED (book: 1, hat: 0, ball: 1)}        \\
        B: DEAL AGREED (book: 1, hat: 1, ball: 1)              \\
    \end{tabular} &
    \begin{tabular}[t]{l}
        A: \bot{i want the hat and 2 books}                          \\
        B: i would like the hat and one book                         \\
        A: \bot{i would need the hat and the books}                  \\
        B: ok sounds fair                                            \\
        A: \bot{DEAL AGREED (book: 2, hat: 1, ball: 0)}        \\
        B: DEAL AGREED (book: 0, hat: 0, ball: 2)              \\
    \end{tabular} \\
\bottomrule
\end{tabular}
} 
\caption{Example human-bot dialogue on \fb{}.
    Bot utterances are bolded.
    \slword{} is quick to concede, while \slstate{} is generally harder to persuade.}
\label{tab:fb-human-bot-chats}
\end{table*}

\end{document}